\documentclass[sigconf]{acmart}

\usepackage{graphicx}
\usepackage{amsmath}
\usepackage{booktabs,makecell, multirow, tabularx}

\usepackage{amssymb}
\AtBeginDocument{%
  }

\copyrightyear{2025}
\acmYear{2025}
\setcopyright{acmlicensed}
\acmConference[MM '25]{Proceedings of the 33rd ACM International Conference on Multimedia}{October 27--31, 2025}{Dublin, Ireland}
\acmBooktitle{Proceedings of the 33rd ACM International Conference on Multimedia (MM '25), October 27--31, 2025, Dublin, Ireland}
\acmDOI{10.1145/3746027.3755624}
\acmISBN{979-8-4007-2035-2/2025/10}

\begin{document}

\title{Boosting Temporal Sentence Grounding via Causal Inference}

\settopmatter{authorsperrow=4}

\author{Kefan Tang}
\email{kftang@stu.xidian.edu.cn}
\affiliation{%
  \department{School of Electronic Engineering,}
  \institution{Xidian University}
  \city{Xi'an}
  \country{China}
}

\author{Lihuo He}
\email{lhhe@xidian.edu.cn}
\authornote{Corresponding author.}
\affiliation{%
  \department{School of Electronic Engineering,}
  \institution{Xidian University}
  \city{Xi'an}
  \country{China}
}

\author{Jisheng Dang}
\email{dangjsh@mail2.sysu.edu.cn}
\affiliation{
  \department{School of Information Science \& Engineering,}
  \institution{Lanzhou University}
  \city{Lanzhou}
  \country{China}
}

\author{Xinbo Gao}
\email{xbgao@mail.xidian.edu.cn}
\affiliation{%
  \department{School of Electronic Engineering,}
  \institution{Xidian University}
  \city{Xi'an}
  \country{China}
}

\renewcommand{\shortauthors}{Kefan Tang, Lihuo He, jisheng Dang, and Xinbo Gao}

\begin{abstract}
Temporal Sentence Grounding (TSG) aims to identify relevant moments in an untrimmed video that semantically correspond to a given textual query. Despite existing studies having made substantial progress, they often overlook the issue of spurious correlations between video and textual queries. These spurious correlations arise from two primary factors: (1) inherent biases in the textual data, such as frequent co-occurrences of specific verbs or phrases,  and (2) the model’s tendency to overfit to salient or repetitive patterns in video content. Such biases mislead the model into associating textual cues with incorrect visual moments, resulting in unreliable predictions and poor generalization to out-of-distribution examples. To overcome these limitations, we propose a novel TSG framework, causal intervention and counterfactual reasoning that utilizes causal inference to eliminate spurious correlations and enhance the model's robustness. Specifically, we first formulate the TSG task from a causal perspective with a structural causal model. Then, to address unobserved confounders reflecting textual biases toward specific verbs or phrases, a textual causal intervention is proposed, utilizing do-calculus to estimate the causal effects. Furthermore, visual counterfactual reasoning is performed by constructing a counterfactual scenario that focuses solely on video features, excluding the query and fused multi-modal features. This allows us to debias the model by isolating and removing the influence of the video from the overall effect. Experiments on public datasets demonstrate the superiority of the proposed method. The code is available at https://github.com/Tangkfan/CICR.
\end{abstract}




\maketitle

\begin{figure}[t]
 \centering
 \includegraphics[width=0.9\columnwidth]{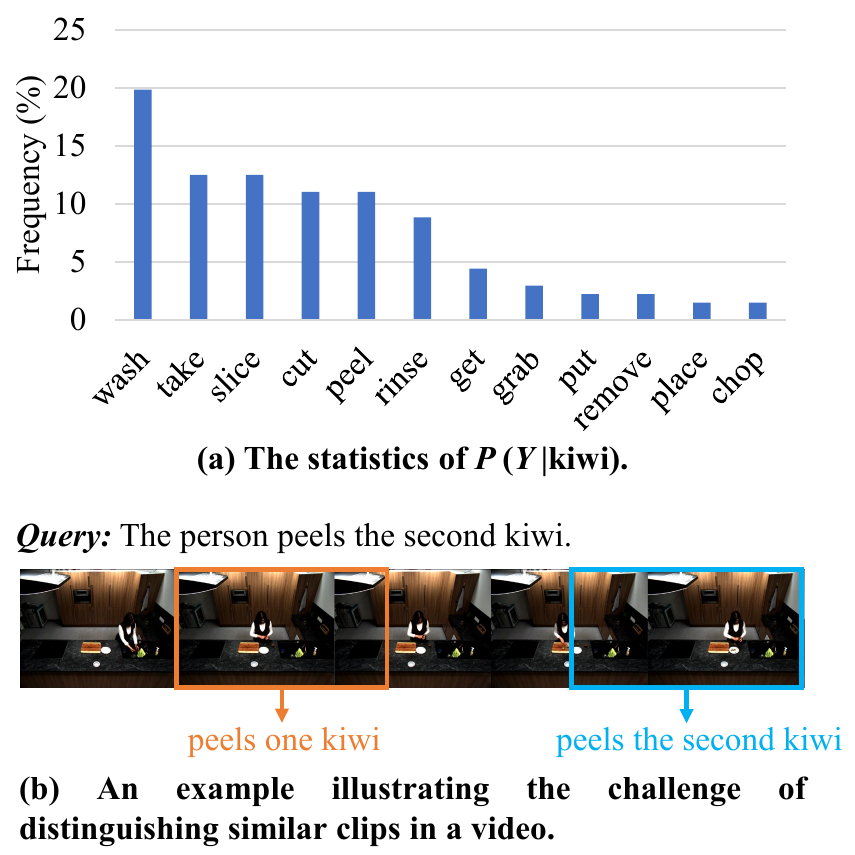} 
 \caption{
 (a) The statistics show that word frequencies differ. This discrepancy can lead the model to become biased, learning high-frequency words effectively while neglecting the meaning of low-frequency words. (b) The clips in the boxes exhibit high semantic similarity, \textit{e.g.}, ``peels one kiwi'' and ``peels the second kiwi''. If the model overly relies on the video content, it may overlook the detailed information in the query. 
}
 \label{motivation}
\end{figure}

\section{Introduction}
Given a natural language query that describes a moment segment in an untrimmed video, Temporal Sentence Grounding (TSG) aims to automatically determine the start and end timestamps of the segment in the video \cite{survey, dsrn}. Although numerous studies on TSG have shown remarkable performance, most of these models remain prone to spurious correlations \cite{ivg, dcm}. This suggests that many TSG models may fail to capture the true correlation between videos and queries. Instead, they often rely on dataset biases or overlook critical information, leading to less reliable predictions. 

As shown in Fig. \ref{motivation}(a), the phrase frequency distribution in the TACoS dataset \cite{tacos} varies significantly. For example, queries involving the object ``kiwi'' often co-occur with actions like ``wash'', ``slice'', and ``chop''. From these frequency statistics, we observe that the token ``kiwi'' is highly correlated with the video moments relevant to ``person washes a kiwi'' but shows a low correlation with ``person chops a kiwi.'' Consequently, a query containing ``person chops the kiwi'' might be inaccurately located in the video segment where ``person washes a kiwi.'' Moreover, video data contains far more information than text \cite{tsmr, Wang_2023_ICCV}, potentially leading the model to over-rely on the video content and overlook the details of queries \cite{mesm}. As illustrated in Fig. \ref{motivation}(b), a single video might include several similar segments, such as ``Person peels the kiwi.'' If the model overlooks specific details in the query, like the temporal sequence ``second'', it may retrieve the wrong segment. This heavy reliance on video data can lead to incorrect matches between queries and video segments, highlighting the importance of models that consider both video and textual inputs equally.

\begin{figure}[t]
 \centering
 \includegraphics[width=0.95\columnwidth]{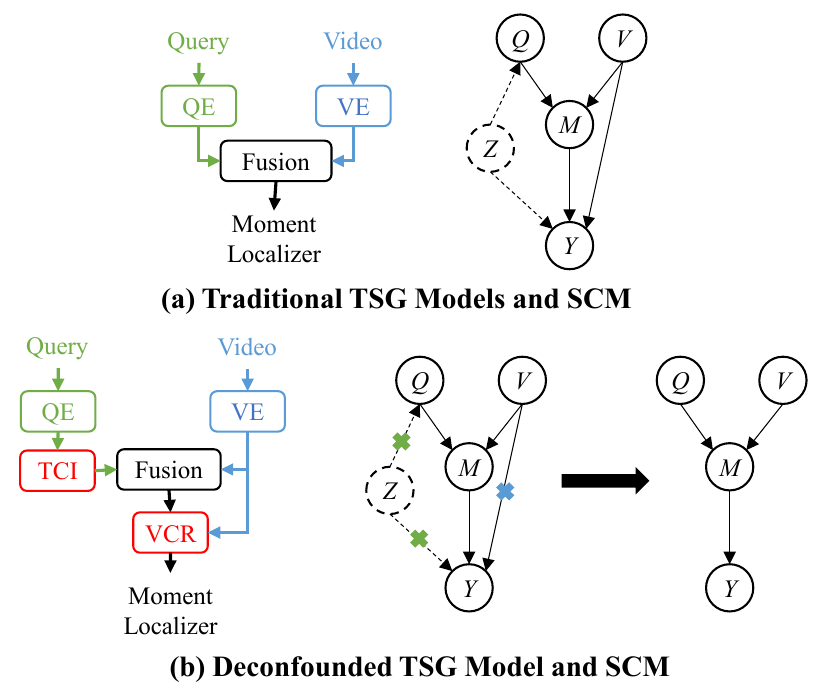} 
 \caption{
 QE: query encoder, VE: video encoder, $V$: video, $Q$: query, $M$: multi-modal features, $Z$: confounder, and $Y$: matching score. 
 (a) Left: The traditional TSG pipeline. 
 Right: The SCM of traditional methods. 
 The confounders $Z$ influence both the query and prediction. Additionally, predictions may be derived solely from video, rather than from multi-modal features.
 (b) Left: Our proposed deconfounded TSG model. 
 Right: The SCM of our approach.
 The TCI blocks the back-door path $Q \leftarrow Z \rightarrow Y$, while the VCR cuts off the direct path $V\rightarrow Y$.
 }
 \label{comparison}
\end{figure}

To explain the spurious correlations, we analyze the TSG model from a causal perspective by utilizing a structured causal model (SCM) \cite{scm}, as shown in Fig. \ref{comparison}. In the SCM, nodes and edges represent variables and causal relationships, respectively. Fig. \ref{comparison}(a) on the left depicts the pipeline of previous methods, where separate encoders process video and query independently, followed by fusion through a modality aligner. These methods are prone to spurious correlations between modalities. A common source of bias in language queries arises from data collectors' preferences for certain actions, potentially leading to unbalanced data representation. Such biases act as unobserved confounders, affecting both queries and predictions, i.e., $Q\leftarrow Z\rightarrow Y$, where $Z$ denotes the confounders, $Q$ represents the query features, and $Y$ indicates the matching score. Additionally, given that video information is more abundant than query data, the model might overly depend on the video alone to determine matching accuracy. This causes the model to over-rely on the direct video path, i.e., $V\rightarrow Y$, resulting in shortcut learning and neglecting multi-modal information. As a result, the model may overlook the query and miss important details, which leads to incorrect retrieval moments. 

To address the aforementioned issues, we propose a novel framework called Causal Intervention and Counterfactual Reasoning (CICR), as shown in Fig. \ref{comparison} (b). This framework aims to eliminate spurious correlations through two key aspects: (1) blocking the backdoor path $Q\leftarrow Z\rightarrow Y$, thus mitigating the impact of unobserved confounders on the query; (2) cutting off the direct path $V \rightarrow Y$, thus ensuring the adequate use of multi-modal information. Specifically, a Textual Causal Intervention (TCI) module is further designed to handle unobserved confounders in the textual modality. This module uses the do-calculus \cite{scm} to calculate the causal effect $P(Y|V,do(Q))$, which differs fundamentally from the traditional likelihood $P(Y|V,Q)$. By employing the do-calculus, the link $Q\leftarrow Z$ is effectively cut off, blocking the backdoor path $Q\leftarrow Z\rightarrow Y$ and eliminating the spurious correlation. As a result, $Q$ and $Y$ are deconfounded, which allows the model to effectively learn the true causal effect. To address the over-dependence on the video modality, we formulate video bias as the direct causal effect on the prediction, which can be removed by subtracting it from the total causal effect. Visual Counterfactual Reasoning (VCR) is proposed to imagine a counterfactual world where only video features $V$ are available, excluding both query features $Q$ and fused multi-modal features $M$. This approach enables us to estimate the video bias by computing the direct causal effect of $V$ on $Y$ and removing it by subtracting it from the total effect on $Y$. 

Our contributions can be summarized as follows:
\begin{itemize}
    \item We analyze the spurious correlations observed in previous TSG methods from a causal perspective by formulating the task within the SCM. This allows us to explicitly model the causal relationship among video, query, and prediction.
    \item We model the annotator's experience bias as an unobserved confounder that introduces query bias. TCI is introduced to mitigate such spurious correlations. Furthermore, VCR is proposed to mitigate negative biases along the direct path $V\rightarrow Y$ while preserving the positive ones.
    \item The proposed method achieves state-of-the-art performance on two public benchmarks. Moreover, the results on the out-of-distribution (OOD) sets of Charades-CD and Charades-CG show its effectiveness in debiasing the base model.
\end{itemize}

\begin{figure*}[ht]
 \centering
 \includegraphics[width=2\columnwidth]{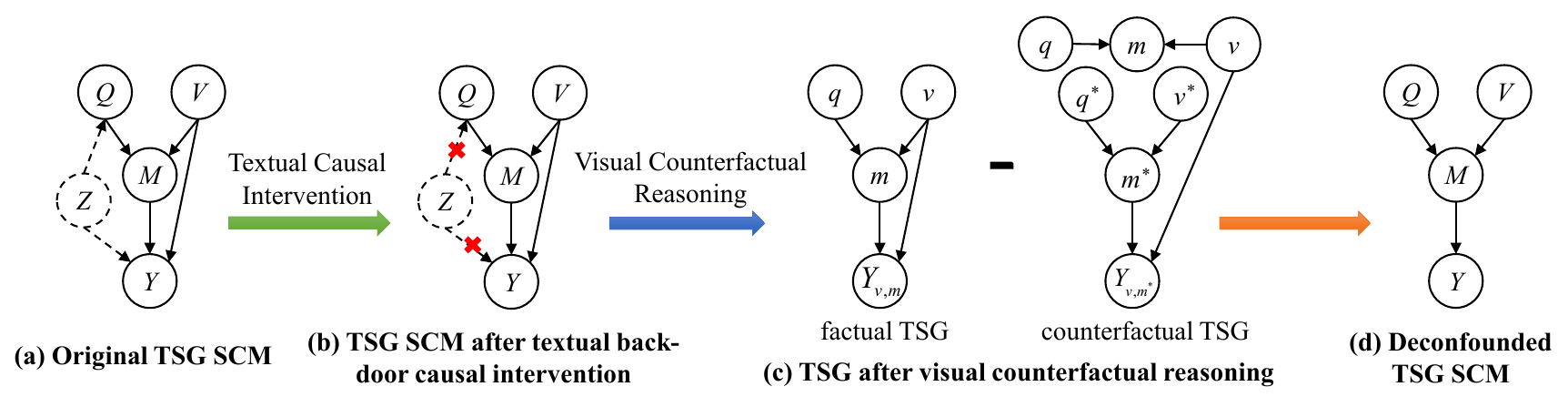} 
 \caption{
The proposed causal graph of CICR. $Q$: query feature. $V$: video feature. $M$: multi-modal feature. $Y$: prediction. $Z$: textual confounder. Uppercase letters (\textit{e.g.}, $V$) denote random variables and lowercase letters (\textit{e.g.}, $v$) denote specific values.}
 \label{causalgraph}
 \end{figure*}
\section{Related Work}
\subsection{Temporal Sentence Grounding}
TSG aims at identifying a specific segment of a video that semantically corresponds to a given language query. Pioneering work can be primarily divided into three categories \cite{m-diff}: proposal-based, proposal-free, and proposal-learnable. The proposal-based methods usually follow a two-step ``proposal-rank'' pipeline. CTRL \cite{ctrl} and MCN \cite{mcn} employ multi-scale sliding windows to generate candidate proposals. To better learn the contextual information in videos, 2D-TAN \cite{2dtan} proposes temporal dependencies between proposals using a two-dimensional temporal map. However, redundant computation is introduced with a large number of negative samples. For more flexibility and efficiency, the proposal-free methods \cite{ablr,lgi,hisa} do not use any proposals and directly regress start and end boundary values or probabilities, which speed advantages but often underperforming. Thus, proposal-learnable methods have been introduced to combine the advantages of both proposal-based and proposal-free approaches. BPNet \cite{bpnet} first generates proposals using a proposal-free backbone, then predicts the final moments. \cite{qvhighlights,qd-detr,mesm} use learnable queries to predict multiple segments. Because proposal-learnable methods adopt either a two-stage prediction or an implicit iterative design, their performance often surpasses that of proposal-free methods. However, existing methods fail to address spurious correlation \cite{ivg,dcm}, resulting in an over-reliance on a single modality while neglecting the true multi-modal information, which ultimately weakens the model's effectiveness and accuracy.

\subsection{Causal Inference}
Causal inference \cite{scm} including causal intervention and counterfactual reasoning has gained widespread application across various areas in the field of computer vision, including video question answer \cite{vcrcnn, cfvqa}, semantic segmentation \cite{conta, demos}, and image captioning \cite{ciic, dic}. It introduces a novel approach to deep learning, offering a way to identify the causal correlations between variables and mitigate spurious correlations. In TSG, DCM \cite{dcm} leverages causal intervention to effectively disentangle and reduce location bias in videos. Simultaneously, IVG-DCL \cite{ivg} applies causal intervention to mitigate biases inherent in language queries. However, both methods focus on a single modality, either video or textual query, and neglect the bias introduced by the other. This one-sided treatment limits their ability to fully address cross-modal spurious correlations. To tackle these issues, we propose CICR, which jointly tackles both textual and video biases through TCI and VCR. CAEM \cite{caem} simulates distribution shift via counterfactual sampling. However, it depends heavily on heuristic data augmentation, while our approach is more straightforward.

\section{Methodology}

\begin{figure*}[h]
\centering
\includegraphics[width=1.95\columnwidth]{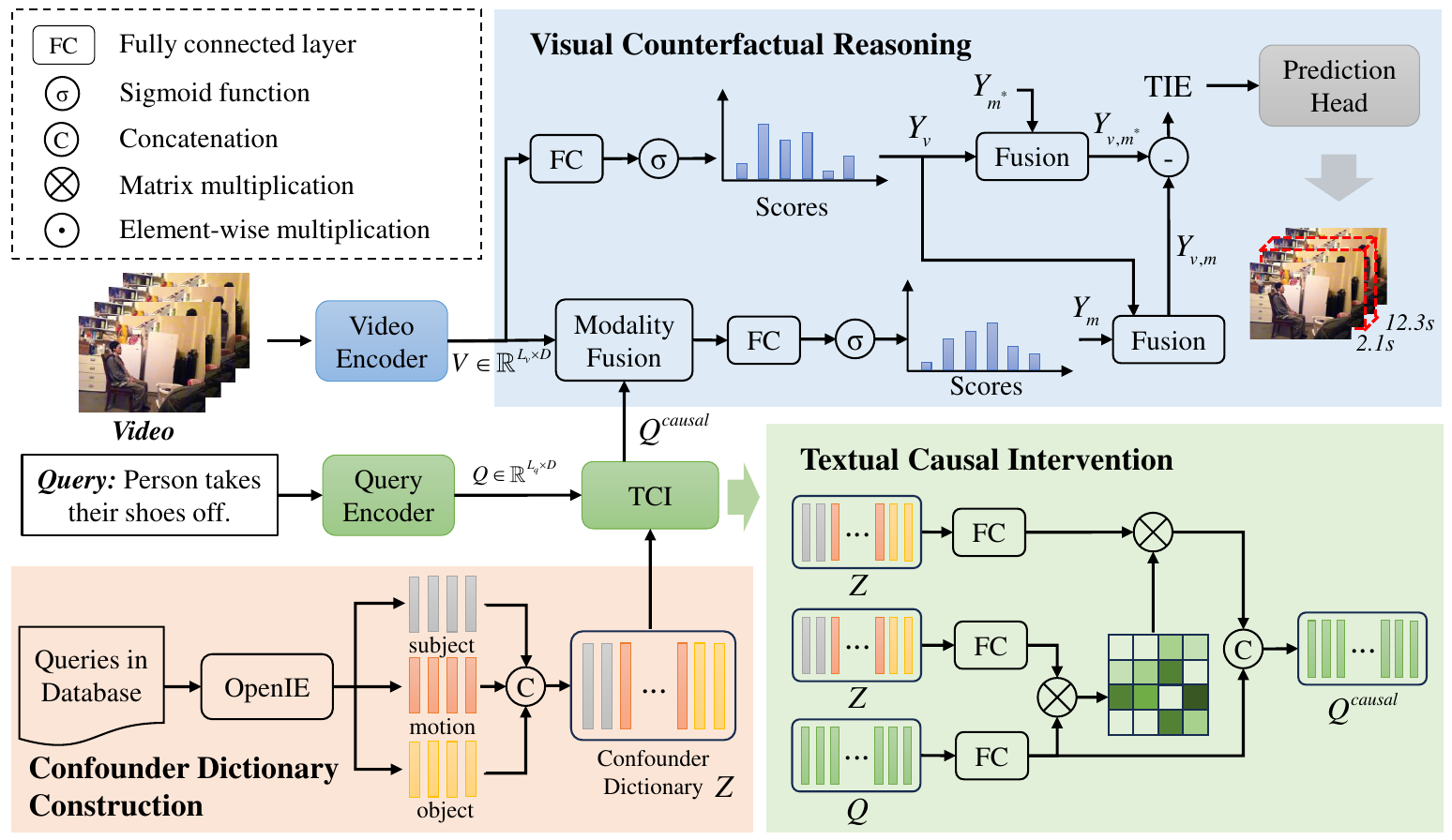} 
\caption{
 Overview of the proposed Causal Intervention and Counterfactual Reasoning (CICR). The model comprises two key components: (1) Textual Causal Intervention (TCI), which mitigates the influence of unobserved confounders by removing non-causal textual biases, and (2) Visual Couterfactual Reasoning (VCR), which blocks the direct path from videos to predictions caused by over-dependece on video.}
\label{model}
\end{figure*}

In this section, the proposed CICR framework is introduced in detail, as illustrated in Fig. \ref{model}. We first define the SCM for TSG, as shown in Fig. \ref{causalgraph}, and derive the causal effect formula. Then, we explain how to estimate the spurious correlations with TCI and VCR. Finally, we introduce the training and inference processes.


\subsection{Causal Graph of TSG} 
To better understand the true causality among variables in TSG models, we use Pearl's SCM \cite{scm}. Traditional TSG methods learn the ambiguous association $P(Y|V, Q)$, ignoring the spurious correlations introduced by textual confounders and the direct effect of videos, as shown in Fig. \ref{causalgraph}. In contrast, our method builds on a causal framework to eliminate these spurious associations, offering a more reliable solution. Below, we detail the rationale behind our causal graph of Fig. \ref{causalgraph}(a).
\begin{itemize}
    \item $Y \leftarrow Z \rightarrow Q \rightarrow M$: The back-door path where confounders $Z$ affect the query $Q$, and ultimately affect the matching score $Y$, leading the model to learn the spurious association. Once the path $Z \nrightarrow Q \rightarrow M \rightarrow Y$ is blocked, $Y$ becomes deconfounded, enabling the model to learn the true causal effect $Q \rightarrow M \rightarrow Y$.
    \item $V \rightarrow Y$: represents a shortcut caused by the video modality, which reflects the negative impact of relying too heavily on video features.
    \item $\{V, Q\} \rightarrow M \rightarrow Y$: represents the indirect effect of $V$ and $Q$ on matching score $Y$ through the fused multi-modal features $M$, which are generated by the modality aligner. The path $V \rightarrow M \rightarrow Y$ captures the positive contribution of video features.
\end{itemize}

\subsection{Feature Extraction}
Following \cite{qvhighlights, mesm}, offline feature extractors are used to get pre-obtained features from the raw data of the video and text. Then, trainable MLPs are utilized to map the extracted video and text feature to a common space. The corresponding video and text features can be represented as $V \in \mathbb{R}^{L_v\times D}$ and $Q  \in \mathbb{R}^{L_q\times D}$, respectively. $K$ is the number of sentences in the video, $D$ is the feature dimension in the common space, and $L_v$ and $L_q$ are the lengths of the video and query features, respectively. 

\subsection{Textual Causal Intervention} 
As shown in Fig. \ref{causalgraph}(b), an unobserved confounder $Z$ introduces spurious correlations in the $Q \rightarrow M \rightarrow Y$ branch by affecting the likelihood $P(Y|V, Q)$. This latent confounding may lead the model to rely on superficial statistical associations rather than true semantic alignment. Traditional TSG models often directly learn the matching score $P(Y|V, Q)$ based on correlations between videos and queries.
The non-interventional prediction can be expressed using the Bayes rule as: 
\begin{equation}
    P (Y|V, Q) = \sum_z P (Y|V, Q, z)P (z|V, Q).
\end{equation}

However, the above objective learns both the main correlation $\{V, Q\} \rightarrow M \rightarrow Y$ and the spurious correlation from the unblocked back-door path $Q \leftarrow Z \rightarrow Y$.
Unlike IVG-DCL \cite{ivg} that performs back-door adjustment \cite{scm} on the fused multi-modal features, our approach conducts deconfounded training along the $Q\rightarrow M \rightarrow Y$ branch before modality fusion. 
This allows the model to obtain cleaner and more reliable textual representations before integrating with visual information. 
Specifically, the do-operator erases incoming arrows into $Q$, blocking non-causal information flow from $Z$ to $Q$. 
In this way, the causal link from $Z\rightarrow Q$ is cut off, isolating the true causal effect $Q \rightarrow Y$ and allowing us to obtain the interventional distribution $P(Y|V, do(Q))$ instead of the biased $P(Y|V, Q)$. The formulations are as follows:
\begin{equation}
\begin{split}
    P(Y|V, do(Q)) 
    &= \sum_z P(Y|V, do(Q), z)P(z|V, do(Q)) \\
    &= \sum_z P(Y|V, Q, z)P(z). 
\end{split}
    \label{P(Y|do(Q))}
\end{equation}

As $Z$ is no longer correlated with $Q$, the causal intervention makes $Q$ fairly incorporate each subject $q$ into the multi-modal fusion $M$, proportionate to its overall distribution.

\noindent\textbf{Confounder Dictionary Construction.}
A typical query in TSG usually consists of a subject, action, and object, aligning with the intuitive structure of ``somebody doing something'' in query descriptions\cite{ivg, cmcir}. To effectively capture these semantic relations, we adopt the Open Information Extraction (OpenIE) model \cite{openie}, which is well suited for extracting verb-centered relation tuples (i.e., subject, action, object) from large-scale natural language data. Based on extracted tuples, we construct three sets of vocabularies: a subject vocabulary $Z_s = \{z_s^1,\ldots, z_s^{L_s}\}$, an action vocabulary $Z_a = \{z_a^1,\ldots,z_a^{L_a}\}$, and an object vocabulary $Z_o = \{z_o^1,\ldots,z_o^{L_o}\}$, where $L_s$, $L_a$, and $L_o$ indicate the respective vocabulary sizes. These components are then integrated into a confounder dictionary consisting of verb-centered relation vocabularies: $Z = [Z_s; Z_a; Z_o] = \{z_1, z_2, \ldots, z_K\}$, where $K=L_s+L_a+L_o$ represents the number of the total types of confounders. This structured representation serves as the foundation for modeling and mitigating textual biases introduced by annotator experience.

\noindent\textbf{Backdoor Adjustment.}
Following \cite{wsvog,CHEN2024102355}, we can adopt the Normalized Weighted Geometric Mean (NWGM) \cite{nwgm} to approximate the deconfounded prediction. The Eq. \ref{P(Y|do(Q))} can be formulated as follow:
\begin{equation}
\begin{aligned}
     P(Y|V, do(Q)) &= \sum_z P(Y|V, \mathcal{G}(Q, z)) P(z) \\
     &\overset{NWGM}\approx P(Y|V, \sum_z\mathcal{G}(Q, z) P(z)),
\end{aligned}
    \label{nwgm}
\end{equation}
where $\mathcal{G}(\cdot)$ is a fusion module that makes the textual embedding of query $Q$ and confounder $z$ fully interact with each other in a common embedding space. Blocking the backdoor path makes $Q$ have a fair opportunity to incorporate every $z$ into $Y$’s prediction, subject to a prior $P(z)$. Eq. \ref{nwgm} is defined as follows:
\begin{equation}
\begin{aligned}
    Q^{causal} &= \sum_z\mathcal{G}(Q, z) P(z) \\
    &= \texttt{Softmax}\left(\frac{(W_kQ)^\top(W_qZ)}{\sqrt{d_k}}\right)(W_vZ),
\end{aligned}
\end{equation}
where $W_k, W_q$, and $W_v$ are learnable parameters to project $Q$ and $Z$ into a joint space. $\sqrt{d_k}$ is a constant scaling factor for feature normalization.

\subsection{Visual Counterfactual Reasoning}

Although we have effectively mitigated textual bias in the $\{V, Q\} \rightarrow M \rightarrow Y$ pathway, the TSG model, as delineated in Fig. \ref{causalgraph}(c), remains susceptible to video-only bias. Based on the proposed causal graph, the final prediction $Y$ is influenced by two branches: the direct effect of the input $V$ on $Y$ through $V \rightarrow Y$, and the indirect effect of both $V$ and $Q$ on $Y$ via the fused features $M$, i.e. $\{V, Q\} \rightarrow M \rightarrow Y$. Formally, the abstract format of the model should be:
\begin{equation}
\begin{aligned}
    Y_{v,m} &= Y(V, M) = \mathcal{H}(Y_v, Y_m)  \\
    &= Y_m \cdot \texttt{Sigmoid}(Y_v),
\end{aligned}
\label{Y(v,m)}
\end{equation}
where $Y_{v,m}$ denotes the prediction under the situation of $M = m$ and $V = v$ and $\mathcal{H}$ denotes the fusion function of $Y_v$ and $Y_m$ to obtain the final score $Y_{v,m}$. Specifically, $Y_v$ and $Y_m$ can be computed as follows:
\begin{equation}
    \begin{aligned}
            Y_v = \mathcal{F}_V(V),\quad     Y_m = \mathcal{F}_M(V,Q),
    \end{aligned}
  \end{equation}
where $\mathcal{F}_V$ is the video-only branch that produces scores based solely on the video and $\mathcal{F}_M$ is the multi-modal branch that fuses two modalities by the modality aligner before computing scores.

As analyzed in Fig. \ref{causalgraph}(c), $Y_{v,m}$ suffers from the negative effect of the direct path $V \rightarrow Y$, which introduces video-only bias. To disentangle this undesirable effect, we leverage VCR to eliminate the spurious influence of $V = v$ and mitigate the model’s over-reliance on video content. Following \cite{bookofwhy}, the total effect (TE) measures how the outcome changes when the input variable shifts from a baseline value to a specific one. In this work, the TE of $V = v$ and $Q = q$ on $Y$ can be calculated by
\begin{equation}
\begin{aligned}
  TE &= Y_{v,m} - Y_{v^*,m^*},  \\
\end{aligned}
\label{te}
\end{equation}
where $v^*$ and $m^*$ denote the reference values of $V$ and $M$, respectively. These reference values correspond to scenarios where the features are unavailable. 

TE can be decomposed into the sum of the natural direct effect (NDE) and the total indirect effect (TIE). Specifically, NDE denotes the effect of $V=v$ on $Y$ along the direct path $V \rightarrow Y$ without considering the indirect effect along $\{V, Q\} \rightarrow M \rightarrow Y$. Thus, the NDE of $V$ on $Y$, which represents the video-only bias, is calculated as follows:
\begin{equation}
\begin{aligned}
    NDE &= Y_{v,m^*} - Y_{v^*,m^*}, \\
\end{aligned}
\end{equation}
where $Y_{v,m^*}$ denotes $V$ set to $v$ in the path $V \rightarrow Y$, representing the model's prediction under the direct effect of the visual modality in the TSG model.

TIE measures the difference between TE and NDE, reflecting how $Y$ changes when $V$ transitions from its reference state (i.e., $V=v^*$) to its actual state (i.e., $V=v$) along the indirect effect path $\{V, Q\} \rightarrow M \rightarrow Y$. To mitigate the bad effect of $V = v$ in the direct path, we subtract NDE from TE to calculate the TIE for two modalities:
\begin{equation}
    \begin{aligned}
        TIE &= TE - NDE = Y_{v,m} - Y_{v,m^*} \\
        &= \mathcal{H}(Y_v, Y_m) - \mathcal{H}(Y_v,Y_{m^*}).
    \end{aligned}
    \label{tie}
\end{equation}
Thus, we utilize the reliable indirect effect (i.e., TIE) for the unbiased prediction by counterfactual inference, as shown in Fig. \ref{causalgraph}(c). Note that $m^*$ denotes the situation where $v$ is not provided, i.e. $V=\varnothing$, which is void. As neural models cannot deal with empty inputs, we assume that the model will randomly guess with equal probability by setting $Y_{m^*}=c$ where $c$ denotes a learnable parameter.

\subsection{Training and Inference}
\textbf{Training.}
An improper choice $Y_{m^*}$ may cause the scales of TE and NDE to differ significantly, making TIE dominated by one of them \cite{cfvqa, clue}. To address this, we leverage Kullback-Leibler (KL) divergence to minimize the difference between $Y_{v,m^*}$ and $Y_{v,m}$, ensuring a more reliable estimation of $Y_{m^*}$:
\begin{equation}
    \mathcal{L}_{kl} = \texttt{KL}(Y_{v,m},Y_{v,m^*}).
\end{equation}
Inspired by \cite{detr,qvhighlights}, the temporal sentence grounding loss consists of three parts:
\begin{equation}
  \mathcal{L}_{tsg} = \lambda_{L1}\|y - \hat{y}\|_1 + \lambda_{iou}\mathcal{L}_{iou}(y, \hat{y}) + \lambda_{ce}\mathcal{L}_{ce},
\end{equation}
where $y$ and $\hat{y}$ are the predicted and ground truth moments, $\lambda_{(L1,iou,ce)}$ are the hyper-parameters, $\mathcal{L}_{iou}$ is the generalized IoU loss \cite{iou}, and $\mathcal{L}_{ce}$ is the cross-entropy loss to classify the foreground or background \cite{detr}.

As a result, the final loss is:
\begin{equation}
  \mathcal{L} = \lambda_{kl}\mathcal{L}_{kl} + \mathcal{L}_{tsg},
\end{equation}
where $\lambda_{kl}$ is the hyper-parameter to balance the two loss terms.
\\
\textbf{Inference.}
In the testing stage, we use the debiased TIE for inference as Eq. \ref{tie}.
\section{Experiments}
To demonstrate the effectiveness of the proposed method, we perform experiments on two public datasets: TACoS \cite{tacos} and QVHighlights \cite{qvhighlights}.
We also experiment on Charades-CD \cite{charades-cd} and Charades-CG \cite{charades-cg}, both of which are OOD variants of Charades-STA \cite{ctrl}.
\subsection{Datasets and Evaluation Metrics}
\noindent\textbf{Datasets.}
1) \textit{TACoS} includes much fewer but longer videos with many more sentences within a video, which is divided into three sets: training, validation, and testing, consisting of 10,146, 4589, and 4083 video-query pairs, respectively.
2) \textit{QVHighlights} contains over 10,000 videos annotated with human-written text queries, serving as a fair benchmark since evaluation on the text split requires submission to the official server\footnote{\href{https://codalab.lisn.upsaclay.fr/competitions/6937}{https://codalab.lisn.upsaclay.fr/competitions/6937}}.
3) \textit{Charades-CD} is divided into four splits: training, validation, test-iid, and test-ood. The first three are independent and identically distributed, while the test-ood set out-of-distribution samples for evaluating debiasing. 
4) \textit{Charades-CG} consists of three testing splits: Test-Trivial, Novel-composition, and Novel-Word. Test-Trivial includes compositions seen during training, Novel-Composition features unseen combinations of known concepts, and Novel-Word contains words that do not appear in the training vocabulary. 

\noindent\textbf{Evaluation Metrics.}
Following previous work \cite{m-diff, mesm}, we adopt R1@$m$, mIoU, mAP@$n$, and mAP$_{avg}$ as evaluation metrics. 
R1@$m$ measures the percentage of instances where the top-1 predicted moment exhibits an Intersection over Union (IoU) with the ground truth (GT) higher than $m$. 
mIoU reflects the average IoU with GT across all test samples.
mAP@$n$ represents the mean average precision with an IoU greater than $n$, while mAP$_{Avg}$ is the average mAP@$n$ over IoU thresholds from 0.5 to 0.95 at 0.05 intervals.

\begin{table}[t]
\caption{Performance comparison with the SOTA methods on the TACoS dataset.}
    \begin{tabular}{ccccc}
    \hline
    Methods & R1@0.1  & R1@0.3  & R1@0.5  & mIoU  \\ \hline
    2D-TAN \cite{2dtan}
            & 47.59 & 37.29 & 25.32 & - \\
    DPIN \cite{dpin}
            & 59.04 & 46.74 & 32.92 & - \\
    FVMR \cite{fvmr}
            & 53.12 & 41.48 & 29.12 & - \\
    IVG-DCL \cite{ivg}
            & 49.36 & 38.84 & 29.07 & 28.26 \\
    VSLNet \cite{vslnet}
            & - & 32.04 & 27.92 & 26.40 \\
    MMN \cite{mmn}
            & 51.39 & 39.24 & 26.17 & - \\
    HCLNet \cite{hclnet}
            & 62.03 & 50.04 & 37.89 & 34.80 \\
    G2L \cite{g2l}
            & - & 42.74 & 30.95 & - \\
    MomentDiff \cite{m-diff}
            & 56.81 & 44.78 & 33.68 & -  \\
    SDN \cite{sdn}
            & 58.56 & 46.14 & 34.84 & 33.32 \\
    MCMN \cite{mcmn}
            & 61.35 & 48.84 & 35.95 & - \\ \hline
    MESM \cite{mesm} (ori.)    & 65.03 & 52.69 & 39.52 & 36.94 \\
    MESM (rep.)
            & \underline{63.93} & \underline{52.96} &\underline{39.27} & \underline{36.44}  \\ 
    CICR (ours) & \textbf{65.23} & \textbf{53.31} & \textbf{39.64} & \textbf{37.53}
  \\ 
    \hline
    \end{tabular}
    
    \label{tacos}
\end{table}

\begin{table}[t]
    \centering
    \caption{Performance comparison with the SOTA methods on the QVHighlights test split.}
    \begin{tabular}{cccccc}
    \hline
    \multirow{2}{*}{Methods} 
    & \multicolumn{2}{c}{R1}   & \multicolumn{3}{c}{mAP}  
    \\ \cmidrule(l){2-3} \cmidrule(l){4-6} 
    & @0.5 & \multicolumn{1}{c}{@0.7} & @0.5 & \multicolumn{1}{c}{@0.75} & \multicolumn{1}{c}{Avg.} \\ \hline
    MCN \cite{mcn}
            & 11.41 & 2.72  & 24.94 & 8.22 & 10.67 \\
    XML \cite{qvhighlights}
            & 41.83 & 30.35 & 44.63 & 31.73 & 32.14 \\
    M-DETR \cite{qvhighlights}
            & 59.78 & 40.33 & 60.51 & 35.36 & 36.14 \\
    UMT \cite{umt}
            & 60.83 & 43.26 & 57.33 & 39.12 & 38.08 \\
    UniVTG \cite{univtg}
            & 58.86 & 40.86 & 57.60 & 35.59 & 35.47 \\
    QD-DETR \cite{qd-detr}
            & 62.40 & 44.98 & 62.52 & 39.88 & 39.86 \\
    MomentDiff \cite{m-diff}
            & 57.42 & 39.66 & 54.02 & 35.73 & 35.95 \\
    VCSJT \cite{vcsjt}
            & 59.14 & 42.02 & 55.76 & 37.79 & 36.37 \\\hline
    MESM \cite{mesm} (ori.)
            & \underline{62.78} & \underline{45.20} & \underline{62.64} & \textbf{41.45} & \underline{40.68} \\ 
    CICR (ours) & \textbf{63.16} & \textbf{45.91} & \textbf{63.85} & \underline{41.39} & \textbf{41.45}
 \\ 
    \hline
    \end{tabular}
    \label{qvhighlights}
\end{table}

\subsection{Implementation Details}
For a fair comparison, we use the publicly available visual and textual features following \cite{mesm, m-diff}. 
Specifically, we use CLIP+SlowFast (C+SF) \cite{clip,sf} as the video feature extractor for QVHighlights and Charades-CG, C3D \cite{c3d} for TACoS, and VGG \cite{vgg} for Charades-CD. For query features, we adopt CLIP for QVHighlights and Charades-CG, and GloVe \cite{glove} for the other two datasets. 
We set $\lambda_{kl}$ as 0.05 for TACoS, and 0.1 for QVHighlights, Charades-CD, and Charades-CG. 
The hidden dimension of the transformer layers is 256. 
We use $\lambda_{L_1} = 10$, $\lambda_{iou} = 1$, and $\lambda_{ce} = 4$ for TSG loss $\mathcal{L}_{tsg}$ as \cite{qvhighlights}. 
Our model is trained with Adam optimizer \cite{adam} with an initial learning rate of 1e-4 and weight decay 1e-4 on a single NVIDIA RTX 3090 with 12 batch size.


\begin{table}[t]
    \centering
    \caption{Performance comparison with the SOTA methods on the Charades-CD OOD set.}
    \begin{tabular}{ccccc}
    \hline
    Methods & R1@0.3  & R1@0.5  & R1@0.7  & mIoU  \\ \hline
    CTRL \cite{ctrl}
            & 44.97 & 30.73 & 11.97 & - \\
    2D-TAN \cite{2dtan}
            & 43.45 & 30.77 & 11.75 & - \\
    VSLNet \cite{vslnet}
            & 48.08 & 32.72 & 19.61 & - \\
    CMNA \cite{cmna}
            & 52.21 & 39.86 & 21.38 & - \\
    DFM \cite{dfm}
            & 56.49 & 41.65 & 23.34 & - \\ 
    BSSARD \cite{bssard}
            & - & 47.20 & 27.17 & 44.59  \\
    CAEM \cite{caem}
            & - & 48.72 & 26.36 & - \\ \hline
    MESM (rep.) 
            & \underline{69.69} & \underline{54.48} & \textbf{29.39} & \underline{47.55}  \\ 
    CICR (ours) & \textbf{70.73} & \textbf{54.72} & \underline{29.30} & \textbf{47.89}
  \\ 
    \hline
    \end{tabular}
    
    \label{charades-cd}
\end{table}
\subsection{Comparisons with the State-of-the-art}
In this section, CICR is compared with the state-of-the-art (SOTA) methods on two datasets: TACoS and QVHighlights. The best result in each case is \textbf{bolded}, while the second-best is \underline{underlined}.

\noindent\textbf{TACoS.} 
We compare the performance of CICR with the SOTA methods, as shown in Tab. \ref{tacos}. The results demonstrate that CICR outperforms the SOTA methods and maintains top positions across all evaluation metrics. Specifically, when compared to the baseline MESM, CICR achieves an improvement of 0.35 in R1@0.3, 0.37 in R1@0.5, and 1.09 in mIoU, respectively. These improvements validate the effectiveness of our method.

\noindent\textbf{QVHighlights.} 
As shown in Table \ref{qvhighlights}, our proposed CICR achieves superior performance compared to the baseline model MESM, with approximately 0.5 improvement across the R1@$n$ metrics. Although mAP@0.75 of CICR is slightly lower than the MESM by 0.06, it is worth noting that CICR surpasses MESM on more representative metric mAP$_{Avg}$, achieving a gain of 0.77. Furthermore, CICR also yields a 1.21 improvement. These results demonstrate the effectiveness of CICR.

\begin{table*}[t]
\centering
\caption{Performance comparison with the SOTA methods on the Charades-CG, which contains two types of OOD settings on Charades-STA: Novel-Composition and Novel Word.
}
\begin{tabular}{ccccccc}
\hline
\multirow{2}{*}{Methods} & \multicolumn{3}{c}{Novel-Composition} & \multicolumn{3}{c}{Novel-Word} \\ 
\cmidrule(l){2-4} \cmidrule(l){5-7} 

& R1@0.5 & R1@0.7 & mIoU  & R1@0.5 & R1@0.7  & mIoU    
\\ \hline
M-DETR      & 37.65 & 18.91 & 36.17 & 43.45 & 21.73 & 38.37    \\
QD-DETR     & 40.62 & 19.96 & 36.64 & 48.20 & 26.19 & 43.22 \\
VDI         &   -   &   -   &   -   & 46.47 & 28.63 & 41.60 \\\hline
MESM (ori.)        & 44.39 & 23.27 & 39.89 & 52.66 & 31.22 & 46.38
\\ 
MESM (rep.)       & \underline{43.64} & \textbf{22.89} & \underline{39.16} & \underline{52.09} & \underline{29.50} & \underline{46.06} \\ 
CICR (ours) & \textbf{44.94} & \underline{22.60} & \textbf{40.25} & \textbf{53.09} & \textbf{31.94} & \textbf{47.13}
\\ \hline
\end{tabular}
\label{charades-cg}
\end{table*}

\subsection{Experiments on Annotation Bias}
As CICR is designed to eliminate spurious correlations between modalities, it enhances the model's ability to generalize and better capture the underlying relationships between videos and language queries. This improved generalization suggests that our method should perform more robustly in rare scenarios. To validate this, we conduct experiments on two OOD datasets: Charades-CD and Charades-CG.

\noindent\textbf{Charades-CD.}
As shown in Tab. \ref{charades-cd}, our proposed CICR achieves the highest performance on most metrics on the Charades-CD. Specifically, CICR achieves an R1@0.3 of 70.73, R1@0.5 of 54.72 and the best mIoU of 47.89, outperforming the previous best baseline MESM by 1.04, 0.24, and 0.34 points, respectively. Although the stricter R1@0.7 score is slightly lower than MESM, the difference 0.09 is negligible. These results demonstrate that CICR exhibits strong generalization ability when encountering unseen or complex query-video patterns in OOD scenarios.

\noindent\textbf{Charades-CG.}
Furthermore, to systematically investigate the model's generalization capacity for infrequent linguistic constructs, we conducted experiments on Charades-CG. The results are shown in Tab. \ref{charades-cg}. Under Novel-Composition split, CICR achieves the best R1@0.5 and mIoU, outperforming both original and reproduced baseline MESM. While the R1@0.7 score is slightly lower than that of MESM, the mIoU improves that the total performance of CICR is better. For the Novel-Word split, CICR shows advantages by achieving the best performance across all metrics, indicating its stronger capability in generalizing to novel linguistic expressions.


\begin{table}[t]
\centering
\caption{Ablation study of different components on QVHighlights val split.}

\begin{tabular}{cccccc}
\hline
\multirow{2}{*}{Methods} 
    & \multicolumn{2}{c}{R1}   & \multicolumn{3}{c}{mAP}  
    \\ \cmidrule(l){2-3} \cmidrule(l){4-6} 
    & @0.5 & \multicolumn{1}{c}{@0.7} & @0.5 & \multicolumn{1}{c}{@0.75} & \multicolumn{1}{c}{Avg.} 
\\ \hline
CICR      &  63.87  &  \textbf{47.35} &  \textbf{64.01}  &  \textbf{41.97} & \textbf{41.98}   \\
w/o TCI   &  63.10  &  46.90 &  63.37  &  41.26 & 40.91   \\ 
w/o VCR   &  \textbf{64.19}  &  47.16 &  63.11  &  41.06 & 41.36   \\ 
w/o both  &  62.26  &  46.45 &  62.05  &  40.61 & 40.62  
\\ \hline
\end{tabular}

\label{abl_comp}
\end{table}

\begin{table}[t]
	\centering
    \caption{Integrating our framework to QD-DETR on QVHighlights val split.}
    
	\begin{tabular}{cccccc}
		\hline
		\multirow{2}{*}{Methods} 
        & \multicolumn{2}{c}{R1}   
        & \multicolumn{3}{c}{mAP}  
        \\ \cmidrule(l){2-3} \cmidrule(l){4-6} 
        & @0.5 & \multicolumn{1}{c}{@0.7} & @0.5 & \multicolumn{1}{c}{@0.75} & \multicolumn{1}{c}{Avg.} \\ \hline 
        QD-DETR (rep.)  &  61.81  &  47.16  &  61.47  & 41.87  & 40.87    \\
		+ Ours    &  \textbf{62.54}  &  \textbf{47.89}  &  \textbf{62.31}  & \textbf{41.90}  & \textbf{41.38}  
		\\ \hline
	\end{tabular}
	\label{abl_other}
\end{table}

\begin{table}[t]
    \centering
     \caption{Efficiency comparisons with the baseline MESM on the QVHighlights dataset.}
    \begin{tabular}{ccc}
    \hline
         Methods&  Params (M)& Epoch Time (sec)\\ \hline
         MESM&  14.13& 73\\
         CICR (ours)&  14.27& 75\\ \hline
    \end{tabular}
   
    \label{computational}
\end{table}

\begin{figure}[t]
    \centering
    \includegraphics[width=\linewidth]{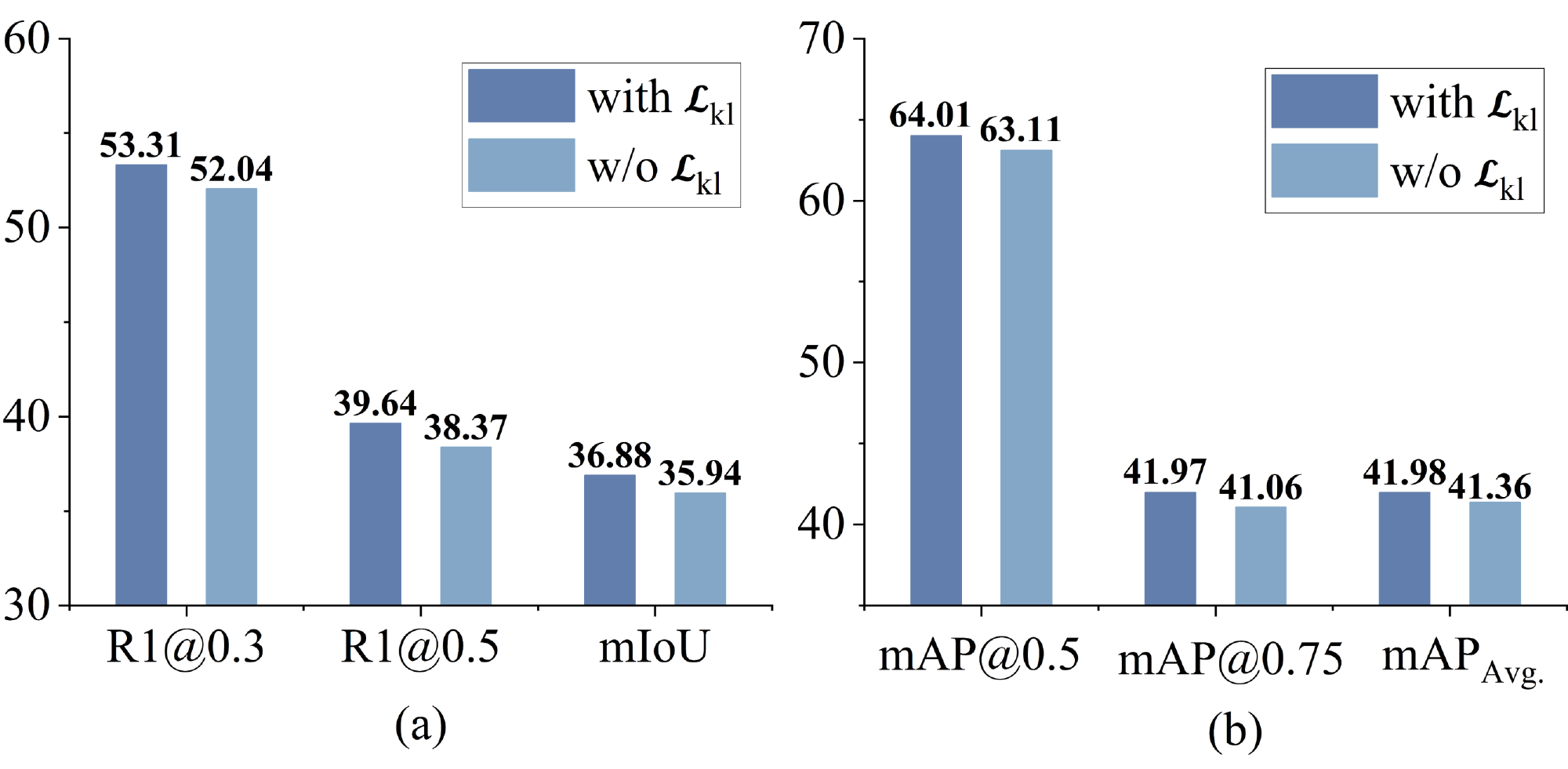}
    \caption{Ablation studies of the impact of loss $\mathcal{L}_{kl}$. (a) on the TACoS dataset. (b) on the QVHighlights val split.}
    \label{abl_loss}
\end{figure}

\begin{figure}[t]
    \centering
    \includegraphics[width=\linewidth]{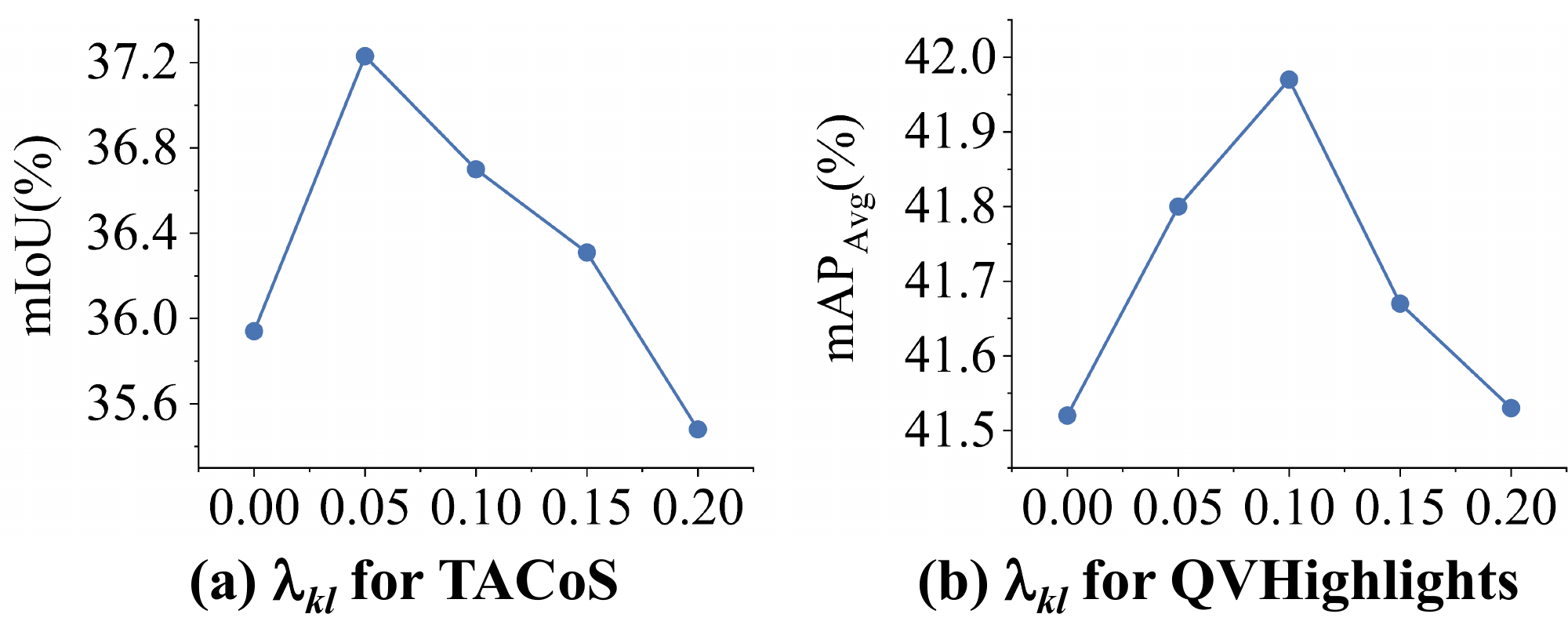}
    \caption{Ablation studies on choices of hyper-parameter $\lambda_{kl}$.}
    \label{hyper-p}
\end{figure}

\subsection{Ablation Study}

\begin{figure*}[t]
    \centering
    \includegraphics[width=\linewidth]{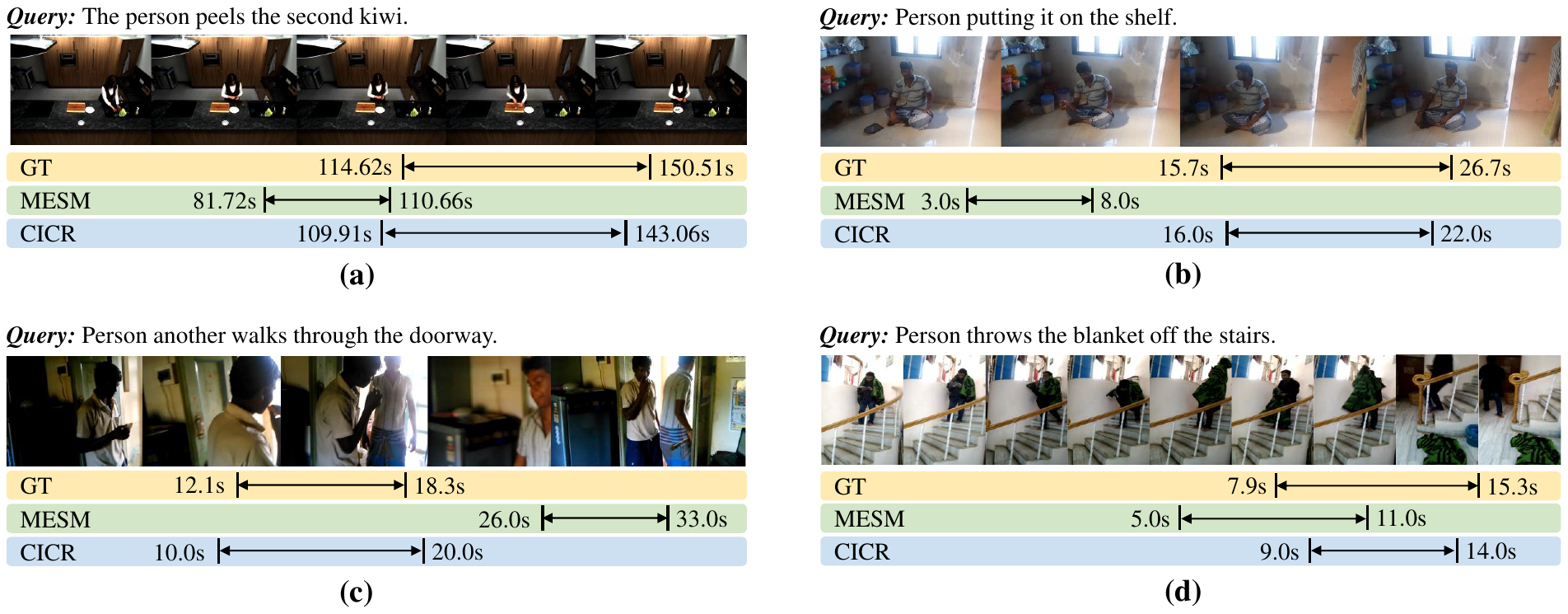}
    \caption{Qualitative examples on three datasets. (a) is from TACoS, (b) from Charades-CG, and (c) and (d) from Charades-CD. Each example contains an input video-query pair, the GT, baseline MESM predictions, and our CICR predictions.}
    \label{visualization}
\end{figure*}

\noindent\textbf{Effectiveness of Each Component.}
We conduct the ablation study of each element in our CICR framework on the QVHighlights val split.
The key components of our CICR are the two causal modules: Textual Causal Intervention (TCI) and Visual Counterfactual Reasoning (VCR). 
As shown in Tab. \ref{abl_comp}, each component is beneficial for TSG.
TCI and VCR achieve gains of 0.74\% and 0.29\% in mAP$_{Avg.}$, respectively. 
When we use all of them, the result comes to the best, which gains 1.36\% in mAP@0.75 and 1.36\% in mAP$_{Avg.}$, demonstrating that both of them are beneficial to the results.

\noindent\textbf{Effectiveness of loss $\mathcal{L}_{kl}$.}
As shown in Fig. \ref{abl_loss}, we conduct an ablation study to investigate the contribution of the KL divergence loss $\mathcal{L}_{kl}$ in CICR. When $\mathcal{L}_{kl}$ is removed, we observe a performance drop across all evaluation metrics. This highlights the crucial role of the KL loss in ensuring the reliable estimation of counterfactual situation $Y_{m^*}$.

\noindent\textbf{Hyper-parameter Sensitive Analysis.}
The hyper-parameter $\lambda_{kl}$ controls the balance of the KL loss $\mathcal{L}_{kl}$ and TSG loss $\mathcal{L}_{tsg}$. In Fig. \ref{hyper-p}, we investigate the impact of varying values of $\lambda_{kl}$ on model performance. It is shown that for the TACoS, the best mIoU is achieved when $\lambda_{kl}=0.05$. For the QVHighlights, the highest mAP$_{Avg}$ is obtained with $\lambda_{kl}=0.1$.

\noindent\textbf{Generality of the method.} 
To validate the generality and flexibility of CICR, we integrate it into QD-DETR \cite{qd-detr} and evaluate on the QVHighlights val split. As shown in Tab. \ref{abl_other}, CICR consistently improves performance across all metrics, demonstrating that our debiasing strategy is not limited to a specific architecture. This result confirms that CICR can serve as a plug-and-play module that effectively enhances various existing temporal grounding models by mitigating modality-induced biases and improving generalization.

\noindent\textbf{Computational complexities.} 
We assess the computational complexity of CICR against the baseline MESM by parameter count and epoch time on the QVHighlights dataset using an NVIDIA RTX 3090 Ti GPU. As shown in Tab. \ref{computational}, CICR adds negligible parameters and maintains comparable epoch time, demonstrating improved performance with minimal overhead.

\subsection{Qualitative Results}
We visualize the localization outcomes of GT, the baseline model MESM, and our proposed CICR on three datasets: TACoS, Charades-CG, and Charades-CD. As illustrated in Fig. \ref{visualization}(a), the keyword ``second'' in the query provides critical temporal information. However, MESM fails to capture this cue due to its over-reliance on video features, resulting in an incorrect prediction. In contrast, our method accurately localizes the target segment, benefiting from the VCR component, which blocks the direct influence of visual bias via the $V \rightarrow Y$ path. This enables the model to better leverage the aligned multi-modal information. In Fig. \ref{visualization}(b), the sample comes from the novel composition split of Charades-CG. While MESM fails to identify the correct moment, our CICR provides a precise prediction, highlighting its improved generalization to previously unseen combinations of familiar concepts. Fig. \ref{visualization}(c) and (d) present two challenging cases from the OOD set of Charades-CD. MESM’s predictions are notably off-target, whereas CICR delivers more accurate temporal grounding. These examples further demonstrate the robustness of TCI in handling distribution shifts and complex scenarios. Overall, these qualitative results clearly illustrate the advantages of CICR in mitigating modality bias and enhancing grounding accuracy across both standard and OOD settings.

\section{Conclusion}
In this paper, a CICR framework is proposed to address the spurious correlations in TSG. Unlike conventional methods, we approach TSG from a causal perspective by constructing the SCM to better understand and address the underlying biases. To alleviate query bias induced by annotators’ experience bias which is modeled as an unobserved confounder, TCI is introduced to block the back-door path $Q \leftarrow Z \rightarrow Y$, thereby eliminating the influence of the confounder $Z$ on the prediction. Additionally, an explainable VCR is proposed to distinguish and reduce over-dependence on video. This enables more balanced multi-modal learning and improves localization precision. Comprehensive experiments on multiple public benchmarks confirm that CICR not only achieves state-of-the-art performance but also exhibits strong generalization to out-of-distribution scenarios.

\begin{acks}
This research was supported partially by the National Natural Science Foundation of China (Grant Nos. 62276203, 62036007) and the Double First-Class Overseas Research Project of Xidian University.
\end{acks}

\bibliographystyle{ACM-Reference-Format}
\bibliography{reference}

\end{document}